# A scalable approach for developing clinical risk prediction applications in different hospitals

Hong Sun[1]*, Kristof Depraetere[1]*, Laurent Meesseman[1], Jos De Roo[1], Martijn Vanbiervliet[1], Jos De Baerdemaeker[1], Herman Muys[1], Vera von Dossow[2], Nikolai Hulde[2], Ralph Szymanowsky[1]*

[1] Dedalus HealthCare, Roderveldlaan 2, 2600 Antwerp, Belgium

{hong.sun, kristof.depraetere, ralph.szymanowsky}@dedalus-group.com

[2] Institute of Anesthesiology, Heart and Diabetes Centre NRW,

Ruhr-University Bochum, Bad Oeynhausen, Germany

{vvondossow, nhulde}@hdz-nrw.de

* Corresponding author

**Objective:** Machine learning (ML) algorithms are now widely used in predicting acute events for clinical applications. While most of such prediction applications are developed to predict the risk of a particular acute event at one hospital, few efforts have been made in extending the developed solutions to other events or to different hospitals. We provide a scalable solution to extend the process of clinical risk prediction model development of multiple diseases and their deployment in different Electronic Health Records (EHR) systems.

**Materials and Methods:** We defined a generic process for clinical risk prediction model development. A calibration tool has been created to automate the model generation process. We applied the model calibration process at four hospitals, and generated risk prediction models for delirium, sepsis and acute kidney injury (AKI) respectively at each of these hospitals.

**Results:** The delirium risk prediction models have on average an area under the receiver-operating characteristic curve (AUROC) of 0.82 at admission and 0.95 at discharge on the test datasets of the four hospitals. The sepsis models have on average an AUROC of 0.88 and 0.95, and the AKI models have on average an AUROC of 0.85 and 0.92, at the day of admission and discharge respectively.

**Discussion:** The scalability discussed in this paper is based on building common data representations (syntactic interoperability) between EHRs stored in different hospitals. Semantic interoperability, a more challenging requirement that different EHRs share the same meaning of data, e.g. a same lab coding system, is not mandated with our approach.

**Conclusions:** Our study describes a method to develop and deploy clinical risk prediction models in a scalable way. We demonstrate its feasibility by developing risk prediction models for three diseases across four hospitals.

# 1. INTRODUCTION

## 1.1. Background and significance

Machine learning (ML) technologies have increased their usage in many industries in the past decade. Driven by the increases of computational power, the advances of machine learning techniques, and the availability of standard electronic health records (EHR), the machine learning algorithms for clinical risk predictions are widely used in healthcare research and applications [1–4]. While much work has been done on developing distinct clinical risk prediction models, few work has been reported on exploration of the scalability of the prediction models, i.e. to extend the risk prediction model development for multiple diseases over different care sites [5]. This paper discusses the scalability issue in clinical risk prediction model development, and presents a scalable approach for prediction model development that is applied on delirium, sepsis and acute kidney injury (AKI) covering four different hospitals.

## 1.2. Scalability of machine learning development on EHR

The main challenge for achieving scalable ML for clinical risk prediction is the lack of a common data representation. In one dimension it hinders the ease of use in different care sites. In the second dimension it hinders the ease of use for different clinical risk predictions. Scaling the development of predictive models is difficult when custom datasets with specific variables are required for different predictive models [4]. Rajkomar et al. [5] achieve both types of scalability by designing a single data structure based on FHIR standard [6], and developing different clinical scenarios over two hospitals with such a common data structure. While the clinical scenarios presented in [5] are mainly administrative scenarios, such as predicting readmissions, this paper presents a scalable approach that is used to develop three clinical risk prediction use cases respectively. This approach is also used to generate clinical risk prediction models in four different hospitals. This paper also describes the steps that are required to develop different clinical risk prediction models, and discusses the necessary interoperability between different EHRs to achieve scalability.

## 1.3. The delirium use case

Delirium is a clinical syndrome defined as an organically caused disturbance in attention and awareness over a short period of time, which has a fluctuating course [7]. While the incidence of delirium after general surgery ranges between 5% and 50% depending on study populations and institutions [8], it is one of the most frequent complications in hospitalized geriatric patients with a prevalence approaching 60% in elderly peoples' and nursing homes [32]. Delirium increases the risk of death during hospitalization [10]. The ability to accurately predict the risk of a patient in developing delirium could enhance screening and prevention efforts. Proactive intervention reduces delirium by over 30% and severe delirium by over 50% [8, 9]. The Confusion Assessment method (CAM) is recommended as a screening tool for delirium [10].

Several machine learning models predicting the risk of delirium have been reported [8, 11, 12], with the area under the receiver-operating characteristic curve (AUROC) ranging from 0.86 to 0.94. Jauk *et al.* implemented and prospectively evaluated their model in a clinical workflow [12].

## 1.4. The sepsis use case

Sepsis is a life-threatening organ dysfunction caused by a dysregulated host response to infection [13]. If not recognized early and managed promptly, it can lead to septic shock, multiple organ failure and death. Identifying those at risk for sepsis and initiating appropriate treatment prior to any clinical manifestations, would therefore have a significant impact on the overall mortality and cost burden of sepsis [14, 15]. The Systemic Inflammatory Response Syndrome (SIRS) criteria were the most widely used screening method, and Sequential Organ Failure Assessment (SOFA) and qSOFA (quick SOFA) are recommended as a replacement of SIRS to provide more accurate screening [13]. These methods utilize tabulation of various patient vital signs and laboratory results to generate risk scores; however, they do not analyse trends in patient data or correlations between measurements.

Several machine learning models predicting the risk of sepsis have been reported that outperform the aforementioned screening methods [14–16]. Islam *et al.* reviewed existing reports on sepsis predictions and compared the performance (to predict sepsis 3 to 4 hours prior to onset) between traditional prediction methods with machine learning approaches: the AUROC with SIRS and SOFA is 0.70 and 0.78 respectively, while seven different studies with machine learning algorithms achieved an overall pooled AUROC of 0.89 [17].

*1.5. The acute kidney injury use case*

Acute kidney injury is a sudden decrease of kidney function which is a common complication that affects as many as one in five hospitalized patients [18, 19]. It is estimated that 30% of hospital-acquired AKI is preventable if predicted in time. AKI is also potentially reversible if diagnosed and managed in time [20]. AKI is diagnosed on the basis of characteristic laboratory findings. The most widely used diagnostic criteria, based on the changes of serum creatinine, have been defined in 2012 by the Kidney Disease Improving Global Outcomes (KDIGO) [21]. However, the elevation of serum creatinine lags behind renal injury, resulting in delayed treatment [22].

Several machine learning models aiming to provide early prediction of AKI have been reported [18, 20, 22]. However, these models do not demonstrate a clinically sufficient level of predictive performance (to predict AKI 24 hours prior to onset), with AUROC less than 0.8. Tomašev *et al.* used a deep neural network to develop an AKI prediction model that demonstrated a very good performance with AUROC of 0.934 (24 hours prior to onset), and 0.921 (48 hours prior to onset) [23]. The main limitation of their work is that the veteran data used to train their model is not general enough, e.g. the female population is relatively small in their dataset.

## 2. OBJECTIVE

The first objective of this study is to develop clinical risk prediction models that can be integrated into a production EHR system in clinical settings. Each model identifies inpatients with a high risk of developing the disease during their hospital stay. The risk prediction is triggered at admission, as well as during the patient stay whenever new observations are available.

The application of machine learning predictions in a clinical setting often faces trust issues because the prediction systems are complex and operate like a 'blackbox' to the end users [2]. Our second objective is to support decision making and increase the confidence of the end users by providing explanations to the predicted results.

The last objective of this study is to introduce a scalable approach for prediction model development and its integration in production EHR systems. A scalable approach would reduce the effort in clinical risk prediction model development by designing a common, reusable model development procedure. Moreover, productizing the development procedure as a toolkit would further reduce the cost to generate prediction models at different care sites based on local data structure.

## 3. MATERIALS AND METHODS

The scalability of developing and deploying different clinical risk prediction models has two aspects:

- First, the scalability to extend the risk prediction model from one disease to another.
- Secondly, the scalability to extend the application of clinical risk prediction from one hospital to another.

This section explains how we reach these two types of scalabilities, as well as the other objectives that were defined in the objective section.

*3.1. A scalable method to develop different clinical risk prediction models*

A common data structure is the prerequisite to develop different clinical risk prediction models in a scalable way. The work presented in [5] designed a single data structure based on FHIR standard, and developed different clinical scenarios with such a common data structure. The scalable clinical risk prediction model development presented in this paper does not rely on a formal standard to represent the source data, rather, it keeps such a requirement on a minimum level: the source data consists of a set of csv (comma-separated values) tables, where the names of tables and columns are predefined to provide corresponding content for features of the prediction models.

A common set of feature groups is fixed as input to train prediction models for different diseases. We use the following feature groups:

- Gender
- Age group
- Admission type
- Department of stay
- History of diagnoses
- Lab results
- Vital signs
- Medications
- Named clinical entities (extracted from clinical notes)

In order to avoid the potential introduction of leaking information, diagnosis codes and procedure codes assigned during patient stay are not used as features. Discharge letters that are only available at discharge stage are also excluded to avoid potential leaking information. The three use cases presented in this paper are all about risks of an acute event, therefore, diagnosis codes assigned in previous patient stays are still used to construct the diagnosis history feature without removing any use case relevant diagnosis.

Figure 1. shows a generic method to develop different clinical risk prediction models. First, common data preparation is applied to the source data, to generate features that a machine learning model can understand. For example, the age at admission is calculated by checking the year difference between birthday and admission day, which is then further binned into age groups. The common data preparation is using all the available data that is

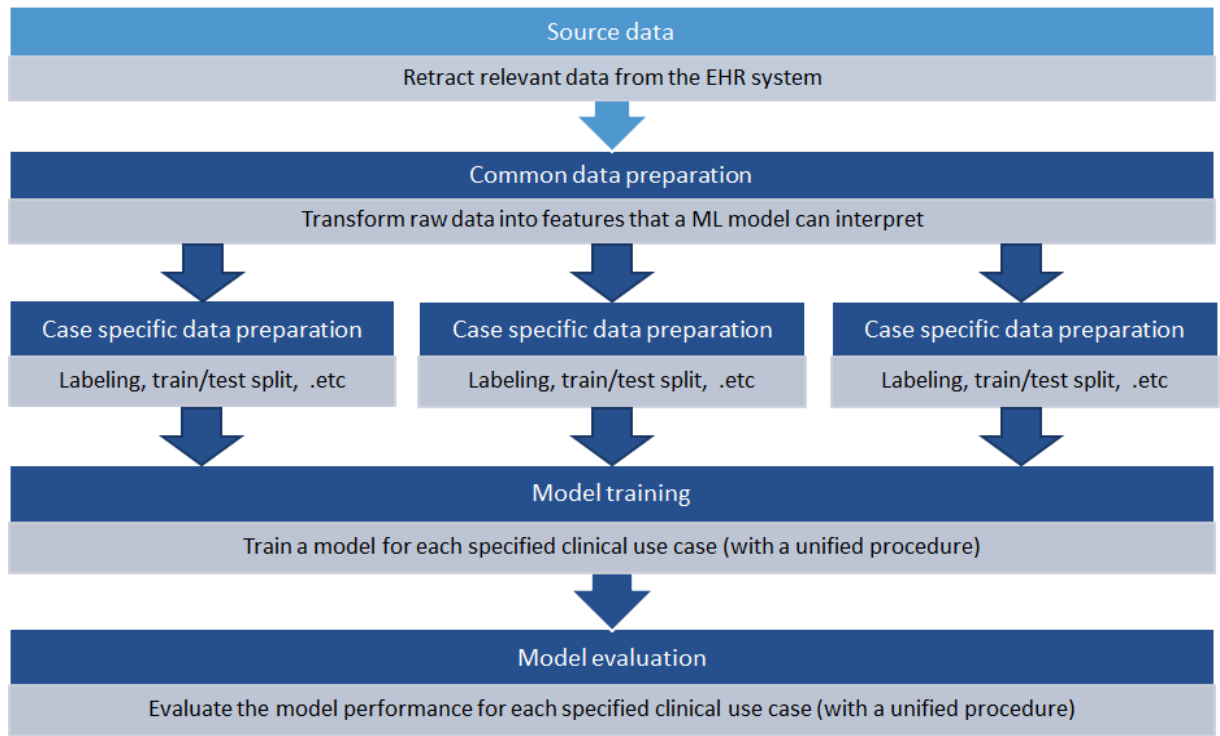

Figure 1. A scalable method to develop risk prediction model for different clinical use cases

generated during a patient stay. In addition, exclusion criteria are applied in order that patients younger than 18 or records with missing basic information are excluded.

Free-text clinical notes, such as admission letters or nursing notes, contain valuable information for clinical risk prediction, and are still not commonly used in most clinical risk prediction models. In [5], clinical notes are used together with other features as inputs of the prediction model, however, when their models are applied at two different sites, the site without clinical notes even delivers better performance than the site providing clinical notes. Rather than feeding the prediction model with the raw clinical notes, we extract named clinical entities from clinical notes, and feed them as features to our prediction models. Clinical entities are extracted from clinical notes using two approaches: in the first approach, we first prepare a set of clinical short phrases that are relevant to the use cases, and then we apply text search to find the phrases in the clinical notes, and return the matched ones as clinical entities; in the second approach, we train a BERT [34] named entity recognition model on German clinical notes, and apply this model on clinical notes to do named entity recognition. Clinical entities extracted by both approaches are merged and used as our clinical entity feature.

Measurements with continuous values, such as lab results, are normalized by assigning an interpretation based on reference values: low, normal or high. For very frequent measurements, such as vital signs, there is an additional normalization based on the number of each interpretation, to indicate the frequency group.

Common data preparation generates a set of features that are ready to be used in model training. However, use case specific data preparation is still needed to prepare the dataset for model training. First, a labeling process assigns the labels for each specified use case. The labeling process is based on the ICD codes assigned to each patient stay at discharge. The list of ICD codes corresponding to each of the three use cases are listed in Appendix A. Secondly, the features prepared in common data preparation may contain data that occurred after the onset of the disease. Such leaking information related to the disease to predict should be removed. The definition of leaking feature varies between different risk prediction use cases. Appendix B provides the definition of leaking features for each of the three use cases. Finally, the labelled data are split patient-wise into train and test and evaluation datasets for model training.

Prediction models for different clinical use cases are trained using the same model training strategy: we use Transformer (Tensor2Tensor) [24] to train a binary classification model for clinical risk prediction. We concatenate the features as inputs, and use the labels as targets for the model training process. We first aggregate all the available information belonging to the same hospital stay to get a complete training record. We also cope with the situation where the model is requested to make predictions in the early stage of a patient stay, when less information is available. For that we apply data augmentation to generate a partial record in combination with the complete record, to enhance the robustness of the clinical risk prediction model. Table 4 in Appendix F shows that with data augmentation the performance of the delirium prediction model is significantly better at the admission time. The training process is repeated to generate a specific prediction model for each clinical risk prediction use case.

Model evaluation for the different use cases is also carried out in a unified procedure by performing predictions on the evaluation data set and doing an evaluation of the predicted results. The metrics such as AUROC, sensitivity, specificity, precision, etc. are calculated to generate an evaluation report. Metrics for different departments are also generated. The importances of different feature groups are also evaluated by checking the impact on the AUROC when a feature group is masked out. Similar to the model training process, the model evaluation process is also repeated to generate a model evaluation for each clinical risk prediction use case.

*3.2. A scalable method to generate clinical risk prediction models at different hospitals*

Ultimately, the goal of developing a clinical risk prediction model is to deploy the developed model to a production system so that risk predictions are made in real time. A prediction model can be developed either at a development site or directly at the target hospital site. Both approaches have certain advantages and disadvantages:

- Developing a prediction model at a development site that uses data of a reference hospital, and deploying the model at another target hospital is convenient for model development. However, it may have degraded performance due to the difference between the data at the development site and the target hospital. Such a degradation would become more severe when there is a difference in vocabulary, e.g. between different lab coding systems.
- Developing a prediction model at a target hospital site produces a model that fits the data characteristics of the target hospital and avoids the aforementioned performance degradation. However, due to privacy concerns, there are often constraints to access the EHR data of a target hospital for model development. In addition, such a process lacks scalability; the model development process needs to be repeated at the different target hospitals.

Our solution combines the advantages of both approaches, and overcomes their disadvantages by carrying out model development at a development site, and applying model calibration at target hospitals. It is therefore convenient to carry out the model development, meanwhile still produce site specific prediction models. Figure 2 shows our scalable method to generate risk prediction models at different hospitals. A development site is selected according to the following criteria: a. The scope and quality of the data for intended use case; b. The site has a special interest in the intended use case; c. The validation process can be actively supported by medical staff of the site; d. The consent of the data protection officer for the use of anonymized patient data for training the model is available.

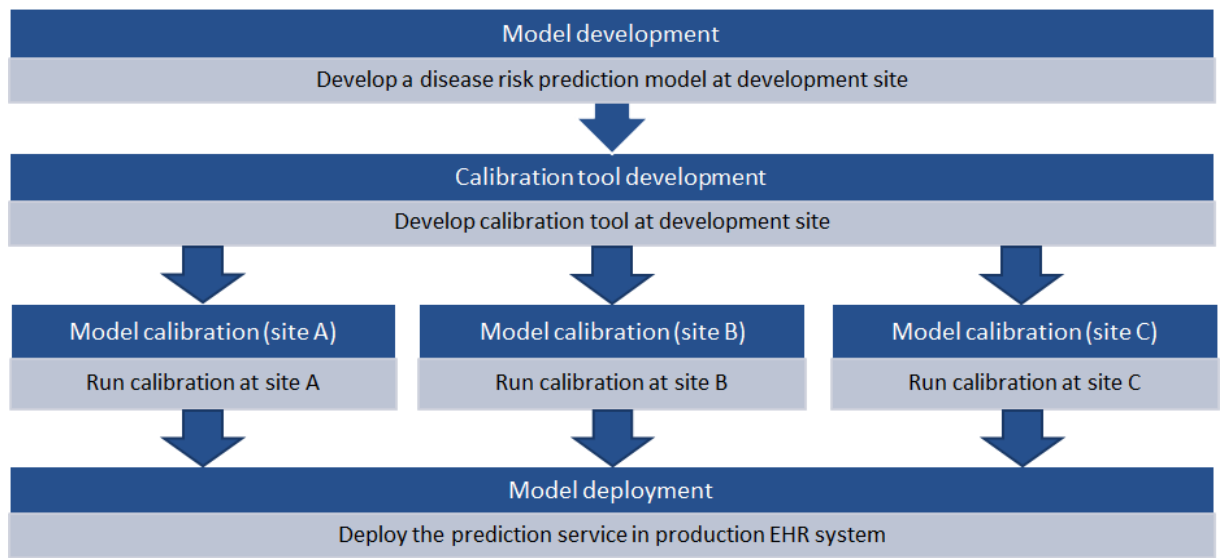

Figure 2. A scalable method to generate risk prediction models at different hospitals

The initial model development is carried out at the development site according to the scalable model development method presented in Figure 1. A toolset is then developed to automate the process of model development. We name this tool the calibration tool where 'calibration' refers to the generation of prediction models at a target hospital, based on their particular data. The calibration tool is delivered as a Docker image. It does not only contain the data preparation and model calibration scripts, but also the needed dependencies such as the required TensorFlow version.

At the target hospital the model calibration process starts with the extraction of source data from the production EHR system. The extracted source data follows the predefined source data formats. Once the source data is extracted, the calibration tool is executed to prepare features for model training, as well as to generate site-specific prediction models. The calibration tool starts with source data analysis to detect abnormalities in the source data. It also automatically adapts, as well as allows manual modification of, the hyperparameters for the model training process. The generated model is trained on the data of the hospital, and therefore considered calibrated with the EHR data of the hospital. The generated models are automatically evaluated, and acceptance criteria are also checked during the model evaluation process.

The generated prediction models that pass the acceptance criteria are used by the prediction service and integrated with the EHR system to provide timely predictions in the production system. The communication between the EHR system and the prediction service, i.e. the prediction request and prediction response, are constructed using the RiskAssessment resource of the FHIR Clinical Module. Other relevant FHIR resources are used to represent the relevant clinical patient data in the risk assessment. Appendix C provides an excerpt of a sample FHIR risk assessment.

### 3.3. Visualized explanations for predicted results

The acceptance of the predictions made by machine learning in clinical practice is often hindered when an explanation of the predicted result is missing. Vaswani [24] and Vig [25] made visualizations of multi-head attentions, i.e. the impact of each token of the input text on each token of the output text. Ribeiro *et al.* introduced the concept of Local Interpretable Model-agnostic Explanations (LIME) to identify the impact of input features to a model over multiple partial representations, that is locally faithful to the classifier [26]. We adopted the concept of LIME to evaluate the impact of each input feature on the predicted result, and visualized the impact of the most influential features as explanations to the predicted result.

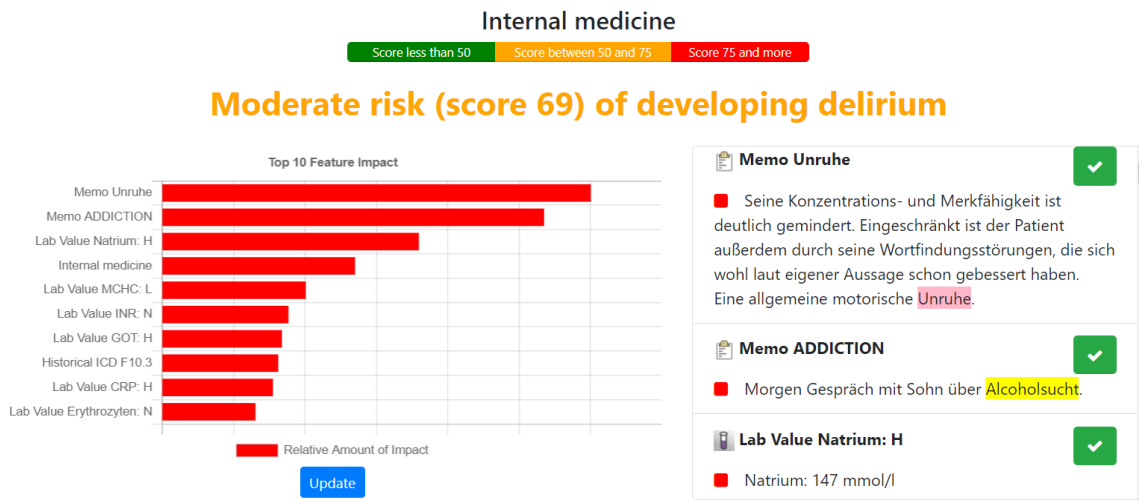

Figure 3. Visualized explanations for predicted result

Figure 3. shows the screenshot of an example of visualized explanation of a sample patient record in German with moderate risk of developing delirium. Based on the risk score, a prediction is classified as low risk (score < 0.5), moderate risk (0.5 ≤ score ≤ 0.75) or high risk (score > 0.75). The left pane of Figure 3 shows the top 10 features that contributed to the prediction of delirium. The features are already normalized after data preparation, e.g. lab results are normalized as high or low, and clinical entities are extracted from documents. The right pane shows the original source of the features displayed in the left pane. We expect the visualized explanations to provide enough details to the physicians, so that they can build confidence in the predicted results.

## 4. RESULTS

### 4.1. Implementations

We started with developing the delirium risk prediction model at our development site. We then developed the sepsis risk prediction model based on the existing scripts for delirium prediction development. The scripts were further adapted following the method presented in Figure 1, to provide a generic approach for risk prediction model development. The calibration tool was developed thereafter and was executed at the target hospitals to generate site specific prediction models for delirium and sepsis respectively. The AKI risk prediction model was later developed following the generic model development approach, and the calibration tool was extended to include the AKI model generation.

Risk prediction models for delirium, sepsis, and AKI were generated by our calibration tool at four different German hospitals: Marienhospital Stuttgart, containing EHR data from 2004 to 2020, HDZ Bad Oeynhausen, containing EHR data from 2009 to 2020, Alexianer Krefeld, containing EHR data from 2014 to 2020, Medius Klinik Nürtingen, containing EHR data from 2009 to 2020. We refer to these four hospitals as Hospital M, Hospital H, Hospital K and Hospital N respectively. The data used for model training and evaluation is anonymized and approved by the data protection officer of each hospital. All four hospitals use the ORBIS® EHR system of Dedalus Group. Except for the clinical notes, the data source for the other features were stored in the same structured tables over these four different sites. The data source extraction script was therefore adapted to cover the different storage of the clinical notes. In addition, lab tests and vital signs are using different coding systems in these four sites, but our calibration tool is designed to be agnostic about the heterogeneity of different coding systems, and does not require manual mapping to a common coding system.

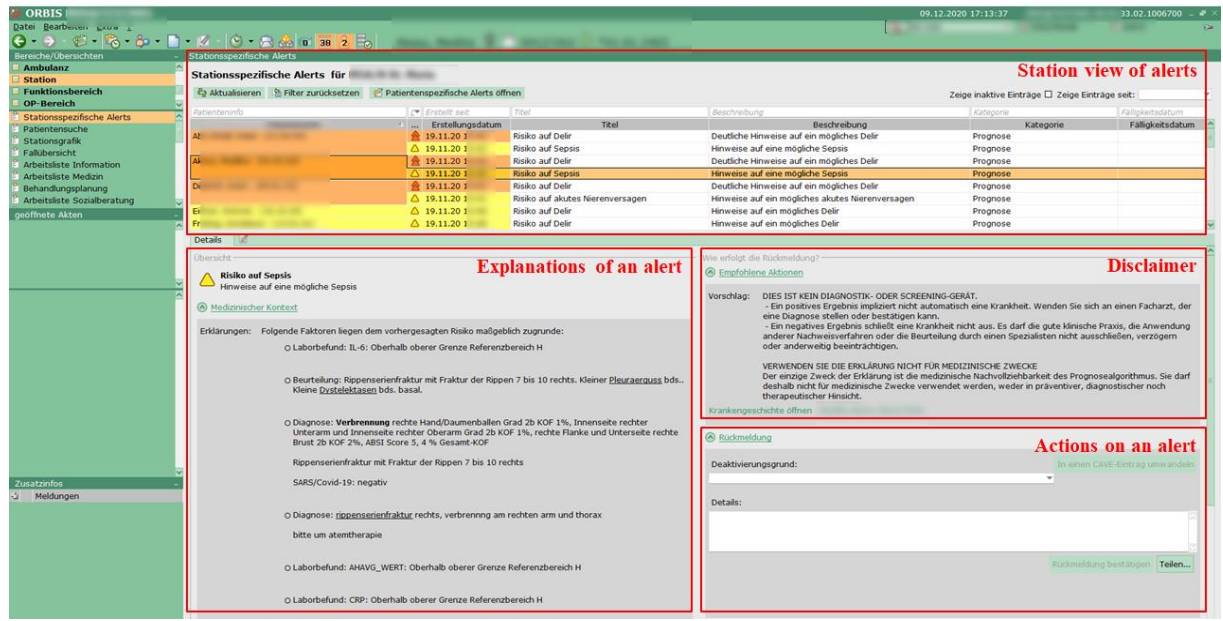

Figure 4. Screenshot of the clinical risk prediction in production EHR system

The target of our clinical risk prediction service is to continuously provide clinical risk predictions during a patient stay when new clinical data is available. The prediction models generated in Hospital M and Hospital H have already been connected to their production EHR system in a controlled setting, to allow stepwise progression and feedback. Figure 4 shows the screenshot of displaying the alerts from the prediction services in a production EHR system. Alerts indicating risks of delirium, sepsis and AKI are displayed in a clinical station view in the upper pane. The alert coloured in yellow indicates a moderate risk ($0.5 \leq \text{score} \leq 0.75$), and the alert coloured in red indicates a high risk (score > 0.75). Explanations of each alert is displayed in the lower left pane, this screenshot shows the example of explanations for a sepsis alert. The treating physician can resolve an alert together with feedback about benefit and (dis)continuation in the right lower pane.

The prediction service integrated with the production EHR system provides timely predictions. In the production system that is not using GPUs and is only using CPUs, the prediction service can process the records of a patient and make a corresponding prediction within one second most of the time. In one of our load tests, we test the performance of our prediction service with 15 threads on a 4 core CPU-only server. In the test, the prediction service reads FHIR risk assessments belonging to a patient (with 81 FHIR resources including 13 clinical notes), processes these FHIR resources, and makes a corresponding prediction: the averaged prediction response time is 261ms for delirium prediction service. The required time to process the records and make a prediction heavily depends on the length of patient records, particularly the size of free text clinical notes. Almost 90% of the data processing time is consumed by processing the free text clinical notes, where a pretrained

TinyBERT model is used to perform named entity recognition. Once the data is processed, our prediction model on average takes less than 30ms to make a prediction.

*4.2. Model performance*

This paper presents the model performance with retrospective data, by the end of the admission day, as well as by the end of the stay, with leaking information excluded. Patients records were randomly split into train (80%) and test (20%) datasets. Table 1 shows the AUROC of the three different prediction models on the test datasets over different stages at four hospitals. The AUROC is lower at the admission day, largely due to the limited data available for prediction at the beginning of the hospital stay. As the stay proceeds, more clinical data becomes available in the EHR system; consequently, the AUROC improves, reaching around 10 percent increase at discharge. The characteristics of the training set at these four hospitals on the three different use cases are enclosed in Appendix D. We also investigated the importance of each feature group in our trained models based on occlusion analysis [27], by checking the impact on the AUROC when a feature group is masked out. The named clinical entities are the most influential feature group in delirium prediction (the impact on AUROC is approximately 8 percent), while the lab results have the greatest impact on sepsis and AKI prediction (the impact on AUROC is approximately 6 percent and 7 percent respectively). A detailed evaluation of the importance of each feature group on the three different use cases over the four sites is provided in Appendix E. We also compared the performance of a delirium prediction model trained in the development site with the delirium prediction model generated with the calibration process at HDZ. Their metrics are compared in Table 4 of Appendix F, it shows the model generated with the calibration process has much better performance, especially at the admission day. Such a comparison confirms the necessity to apply calibration at the deployment site. Nevertheless, the model trained at the development site still achieved fair results at the deployment site at the day of discharge, this provides the possibility to investigate possible improvement with transfer learning, that is to start the calibration process based on the model trained at the development site, rather than from scratch.

Table 1. Area under the receiver-operating characteristic curve (AUROC) with 95% confidence intervals (CI), over different stages of hospital stay on the test datasets of the four different hospitals, for three different clinical use cases

| Hospital | DELIRIUM | | SEPSIS | | AKI | |
|---|---|---|---|---|---|---|
| | admission AUROC (CI 95%) | discharge AUROC (CI 95%) | admission AUROC (CI 95%) | discharge AUROC (CI 95%) | admission AUROC (CI 95%) | discharge AUROC (CI 95%) |
| M | 84.57% [84.3 - 84.8] | 96.74% [96.6 - 96.9] | 87.14% [86.9 - 87.4] | 96.48% [96.3 - 96.6] | 87.20% [87.0 - 87.4] | 93.41% [93.2 - 93.6] |
| H | 75.20% [74.7 - 75.7] | 98.03% [97.9 - 98.2] | 82.67% [82.3 - 83.1] | 96.17% [96.0 - 96.4] | 80.37% [79.9 - 80.8] | 95.75% [95.5 - 96.0] |
| K | 83.19% [82.7 - 83.6] | 92.57% [92.2 - 92.9] | 85.39% [85.0 - 85.8] | 93.91% [93.6 - 94.2] | 83.90% [83.5 - 84.3] | 89.18% [88.8 - 89.6] |
| N | 85.26% [85.0 - 85.5] | 93.09% [92.9 - 93.2] | 89.32% [89.1 - 89.5] | 95.55% [95.4 - 95.7] | 83.65% [83.4 - 83.9] | 88.09% [87.9 - 88.3] |
| avg | 82.05% | 95.11% | 88.23% | 95.53% | 85.43% | 91.60% |

It is observed that the difference in AUROC between admission and discharge is larger in hospital H compared with the other three hospitals, particularly in the delirium use case. This is due to the fact that hospital H is specialized in heart disease and diabetes, with a much larger proportion of heart surgeries. A patient stay in hospital H is on average twice longer than the other three hospitals. Consequently, the gap between averaged length of patient record at discharge and at admission in hospital H is much larger, particularly for textual information such as clinical notes. Since the named clinical entity from clinical notes is the most influential feature in the delirium use case, the lack of clinical notes data at the admission day in hospital H decreases the AUROC.

The AUROC in hospital K is lower compared with the other three hospitals. Hospital K is a smaller hospital specialized in psychiatry. Information for training was limited. First, without intensive care, which is associated with high rates of delirium, sepsis and AKI, the number of positive cases for training was significantly reduced. Secondly, the dataset did not include medication or vital signs, as these had not been recorded in the EHR system. Thirdly, only 6 years of patient records could be provided, which is relatively small compared with other datasets. Nevertheless, even with these limitations, the prediction model in hospital K still has satisfactory results. This shows the presented scalable approach is capable of developing meaningful prediction models when limited data is available.

## 5. DISCUSSION

### 5.1. Interoperability

The scalable approach for developing clinical risk prediction models presented in this paper is relying on data interoperability: a common representation of source data between different hospitals and a common representation of features for different prediction models. There are different levels of clinical data interoperability: syntactic interoperability, that guarantees the smooth data transition between different EHR system by representing data with a common structure; and semantic interoperability that guarantees the common understanding between different EHR systems by representing data with a common semantic [28, 29]. While the syntactic interoperability is easier to reach by representing data in a common structure, the semantic interoperability is a much more challenging task. It requires that clinical data in different EHR systems share the same meaning, by using common coding systems, and clinical terminology mapping between different coding systems is needed [30].

Reaching semantic interoperability could benefit the clinical risk prediction model development. A model trained at a development site could understand the input features from a target hospital to make a prediction, and therefore be deployed directly. In addition, once the features in the development site and the target hospital share the same semantics, it is then more efficient to apply transfer learning: after developing a prediction model using data at the development site, continue with model training with training data from the target hospital to enhance the model performance at that target hospital. This is helpful for target sites that have limited training data [5]. Although reaching semantic interoperability has the aforementioned benefits, according to our previous experience [28, 30], it also comes with substantial costs to create the required semantic mappings. With the target of developing a scalable approach with affordable cost, we put minimum requirements for interoperability.

We require the source data for training to be prepared as csv files with predefined names of tables and columns. This constitutes a minimum level of syntactic interoperability to cover for data structure differences over different hospitals. We use such simple common csv files to format the source data, rather than popular clinical data standards such as FHIR[6] or OMOP [31], for the sake of reducing the cost of source data preparation. Mapping source data to a clinical data standard requires much work, and such an effort is repeatedly required when developing clinical risk predictions at a new hospital. With our approach, source data is extracted and represented with the predefined format of csv tables. It can be easily implemented with SQL scripts or other similar methods, and adapted to the data structure of each target hospital.

While the source data is represented with simple common csv files, we still rely on clinical standards as the interface to our prediction service. We use the RiskAssessment resource of the FHIR Clinical Module to construct the prediction request and response, as a means to communicate between the EHR system and the prediction service. The FHIR resources are used to represent relevant clinical data. Using a standard messaging format improves the reusability of the prediction services over different EHR systems. Since FHIR has gained in popularity and is increasingly adopted by the healthcare industry [33], relying on FHIR messaging to communicate between EHR systems and our prediction service is considered as the most efficient, future-proof solution.

*5.2. Open choices of implementations*

This paper intends to present a scalable approach for risk prediction development, as well as the deployment at different hospitals. We take the use case of delirium, sepsis and AKI to demonstrate the scalability of our approach. Therefore, we consider the choices that we made in our implementation are among many possible solutions. For example, in the delirium risk prediction, Gonzalvo [9] excluded patient with delirium at admission, Kim [8] considered history of prior delirium as a predictor of delirium, and Jauk [12] assigns the highest risk score to patient with a history of delirium. We neither exclude any historical diagnosis, nor exclude any patient with a history of the disease to predict because all these three use cases are predicting the risk of an acute disease. We evaluated the impact of each feature group in each prediction model, and summarized the results in Table 3 of Appendix E. This table shows many of such choices, e.g. the inclusion vs. exclusion of use case related historical diagnosis, will not bring much impact to the metrics of the final model because the corresponding feature group has a low impact on the trained model. In the meantime, we are also continuously improving our solutions by optimizing our implementation choices. The current architecture of our prediction model, together with the hyper parameters are provided in Appendix G.

*5.3. Limitations*

The presented scalable model development approach relies on common feature preparation and similar labeling strategy. The common features that we prepared were sufficient for the three risk prediction use cases presented in this paper. Nevertheless, we also foresee that additional feature processing will be required for other more complex clinical risk prediction use cases. Similarly, more precise labeling urges for more sophisticated labeling methods than ICD codes only. In principle, such extensions can be made by extending the common feature preparation, or applying use case related adaptations. Although our approach is scalable, it should not be considered as being generic to develop prediction models for every disease; rather, it is meant as being generic to develop a set of clinical risk prediction models with similar behavior.

The scalable approach for developing clinical risk prediction models presented in this study was implemented in four different hospitals that were all using ORBIS® EHR systems. It was not yet tested in hospitals with different

EHR systems. However, since only a minimum level of interoperability is expected, we consider our calibration process as feasible across various EHR systems.

The results presented in this paper are based on retrospective data. Our prediction models are now connected to two production EHR systems. A thorough evaluation of the prediction models in these production systems is now ongoing and the outcome will be delivered in a separate paper.

## 6. CONCLUSIONS

Clinical risk prediction applications built with machine learning technologies are now widely used in clinical research and applications. Most of the clinical risk prediction applications are dedicated to predicting the risk of one disease, and the development process is often based on the characteristics of the patient records retrieved from a single hospital. Developing a clinical risk prediction model is often a time-consuming task that involves feature engineering, model training and model evaluation. Developing a clinical risk prediction model using deep learning technology based on a single site is more vulnerable to overfitting. Since such a prediction system is often complex and operates like a 'blackbox', it makes biases, such as overfitting, difficult to detect. In addition, a clinical risk prediction model that is over-specialized on a single site lacks the scalability to be applied on other hospitals, which is a waste of development efforts.

We present a scalable approach for developing, calibrating and deploying clinical risk prediction models in different hospitals, so that the efforts in model development can be reused. Such an approach reduces the risk of overfitting by applying a generic development procedure across different sites. It also aims to build trust from physicians by introducing visualized explanations for predictions. The model development process is reused in developing prediction models for different diseases by sharing a set of common features for predictions, as well as a common strategy for labeling and model training. The model generation and deployment in different hospitals are relying on our calibration tool, in combination with a minimal level of requirements on data interoperability. The presented scalable approach has been used to develop delirium, sepsis, and AKI risk prediction services, and it is used to generate and deploy prediction models at four different German hospitals. While detailed evaluation of the models in clinical settings is still ongoing, this paper presents performance evaluation with retrospective data, which shows satisfactory and promising results.

## 7. AUTHOR CONTRIBUTIONS

KD, MV and HS developed the solution architecture for scalable clinical risk prediction model development. LM and RS defined the clinical use cases. RS performed data extraction. HS, MV developed the data analysis and data preparation. JDR and HS implemented the machine learning algorithm. HM, HS and LM implemented NLP algorithms for named clinical entity recognition. LM, HS, JDR developed model evaluation algorithms. MV and HS developed the model calibration tool. KD and JDB developed the solution architecture for the prediction service. JDB developed the API and GUI for the prediction services. LM, KD, HS, JDR evaluated the model performance. VVD, NH and LM made clinical evaluations of the predicted results. All authors critically revised the manuscript and approved its final version.

## 8. ACKNOWLEDGEMENT

The authors would like to acknowledge Elena Albu for her contribution in model evaluation and detailed review of the paper, Dieter Vanden Abeele and Nico Lapauw for the integration of the prediction services to ORBIS® EHR system, Zuzanna Kwade for reviewing the paper, and Marienhospital Stuttgart, HDZ Bad Oeynhausen, Alexianer Krefeld, and Medius Klinik Nürtingen for assisting the model calibration and prediction model evaluation. Prof. von Dossow and Dr. Hulde would also like to acknowledge the ARGUS project (100126059) from Ruhr-Universität Bochum.

## REFERENCE

[1] Esteva, A., Robicquet, A., Ramsundar, B., Kuleshov, V., DePristo, M., Chou, K., ... & Dean, J. (2019). A guide to deep learning in healthcare. Nature medicine, 25(1), 24-29.

[2] Cutillo, C. M., Sharma, K. R., Foschini, L., Kundu, S., Mackintosh, M., & Mandl, K. D. (2020). Machine intelligence in healthcare—perspectives on trustworthiness, explainability, usability, and transparency. NPJ Digital Medicine, 3(1), 1-5.

[3] Rajkomar, A., Dean, J., & Kohane, I. (2019). Machine learning in medicine. New England Journal of Medicine, 380(14), 1347-1358.

[4] Goldstein, B. A., Navar, A. M., Pencina, M. J., & Ioannidis, J. (2017). Opportunities and challenges in developing risk prediction models with electronic health records data: a systematic review. Journal of the American Medical Informatics Association, 24(1), 198-208.

[5] Rajkomar, A., Oren, E., Chen, K., Dai, A. M., Hajaj, N., Hardt, M., ... & Dean, J. (2018). Scalable and accurate deep learning with electronic health records. NPJ Digital Medicine, 1(1), 18.

[6] Mandel, J. C., Kreda, D. A., Mandl, K. D., Kohane, I. S., & Ramoni, R. B. (2016). SMART on FHIR: a standards-based, interoperable apps platform for electronic health records. Journal of the American Medical Informatics Association, 23(5), 899-908.

[7] American Psychiatric Association. (2013). Diagnostic and statistical manual of mental disorders (DSM-5®). American Psychiatric Pub.

[8] Kim, M. Y., Park, U. J., Kim, H. T., & Cho, W. H. (2016). DELirium prediction based on hospital information (Delphi) in general surgery patients. Medicine, 95(12).

[9] Gonzalvo, C. M., de Wit, H. A., van Oijen, B. P., Deben, D. S., Hurkens, K. P., Mulder, W. J., ... & van der Kuy, P. H. M. (2017). Validation of an automated delirium prediction model (DElirium MOdel (DEMO)): an observational study. BMJ open, 7(11), e016654.

[10] Inouye, S. K., van Dyck, C. H., Alessi, C. A., Balkin, S., Siegal, A. P., & Horwitz, R. I. (1990). Clarifying confusion: the confusion assessment method: a new method for detection of delirium. Annals of internal medicine, 113(12), 941-948.

[11] Wong, A., Young, A. T., Liang, A. S., Gonzales, R., Douglas, V. C., & Hadley, D. (2018). Development and validation of an electronic health record–based machine learning model to estimate delirium risk in newly hospitalized patients without known cognitive impairment. JAMA network open, 1(4), e181018-e181018.

[12] Jauk, S., Kramer, D., Großauer, B., Rienmüller, S., Avian, A., Berghold, A., ... & Schulz, S. (2020). Risk prediction of delirium in hospitalized patients using machine learning: An implementation and prospective evaluation study. Journal of the American Medical Informatics Association, 27(9), 1383-1392.

[13] Singer, M., Deutschman, C. S., Seymour, C. W., Shankar-Hari, M., Annane, D., Bauer, M., ... & Hotchkiss, R. S. (2016). The third international consensus definitions for sepsis and septic shock (Sepsis-3). Jama, 315(8), 801-810.

[14] Nemati, S., Holder, A., Razmi, F., Stanley, M. D., Clifford, G. D., & Buchman, T. G. (2018). An interpretable machine learning model for accurate prediction of sepsis in the ICU. Critical care medicine, 46(4), 547.

[15] Mao, Q., Jay, M., Hoffman, J. L., Calvert, J., Barton, C., Shimabukuro, D., ... & Zhou, Y. (2018). Multicentre validation of a sepsis prediction algorithm using only vital sign data in the emergency department, general ward and ICU. BMJ open, 8(1).

[16] Desautels, T., Calvert, J., Hoffman, J., Jay, M., Kerem, Y., Shieh, L., ... & Wales, D. J. (2016). Prediction of sepsis in the intensive care unit with minimal electronic health record data: a machine learning approach. JMIR medical informatics, 4(3), e28.

[17] Islam, M. M., Nasrin, T., Walther, B. A., Wu, C. C., Yang, H. C., & Li, Y. C. (2019). Prediction of sepsis patients using machine learning approach: a meta-analysis. Computer methods and programs in biomedicine, 170, 1-9.

[18] Hodgson, L. E., Sarnowski, A., Roderick, P. J., Dimitrov, B. D., Venn, R. M., & Forni, L. G. (2017). Systematic review of prognostic prediction models for acute kidney injury (AKI) in general hospital populations. BMJ open, 7(9), e016591.

[19] Silver, S. A., & Chertow, G. M. (2017). The economic consequences of acute kidney injury. Nephron, 137(4), 297-301.

[20] Kate, R. J., Pearce, N., Mazumdar, D., & Nilakantan, V. (2020). A continual prediction model for inpatient acute kidney injury. Computers in biology and medicine, 116, 103580.

[21] Khwaja, A. (2012). KDIGO clinical practice guidelines for acute kidney injury. Nephron Clinical Practice, 120(4), c179-c184.

[22] Mohamadlou, H., Lynn-Palevsky, A., Barton, C., Chettipally, U., Shieh, L., Calvert, J., ... & Das, R. (2018). Prediction of acute kidney injury with a machine learning algorithm using electronic health record data. Canadian journal of kidney health and disease, 5, 2054358118776326.

[23] Tomašev N, Glorot X, Rae J W, et al. A clinically applicable approach to continuous prediction of future acute kidney injury[J]. Nature, 2019, 572(7767): 116-119.

[24] Vaswani, A., Shazeer, N., Parmar, N., Uszkoreit, J., Jones, L., Gomez, A. N., ... & Polosukhin, I. (2017). Attention is all you need. In Advances in neural information processing systems (pp. 5998-6008).

[25] Vig, J. (2019). A Multiscale Visualization of Attention in the Transformer Model. ACL 2019, 37.

[26] Ribeiro, M. T., Singh, S., & Guestrin, C. (2016, August). "Why should I trust you?" Explaining the predictions of any classifier. In Proceedings of the 22nd ACM SIGKDD international conference on knowledge discovery and data mining (pp. 1135-1144).

[27] Zeiler, M. D., & Fergus, R. (2014). Visualizing and understanding convolutional networks. In European conference on computer vision (pp. 818-833). Springer, Cham.

[28] Sun, H., Depraetere, K., De Roo, J., Mels, G., De Vloed, B., Twagirumukiza, M., & Colaert, D. (2015). Semantic processing of EHR data for clinical research. Journal of biomedical informatics, 58, 247-259.

[29] Bhartiya, S., Mehrotra, D., & Girdhar, A. (2016). Issues in achieving complete interoperability while sharing electronic health records. Procedia Computer Science, 78(C), 192-198.

[30] Hussain, S., Sun, H., Sinaci, A. A., Erturkmen, G. B. L., Mead, C. N., Gray, A. J., ... & Forsberg, K. (2014). A framework for evaluating and utilizing medical terminology mappings. In MIE (pp. 594-598).

[31] OHDSI. OMOP common data model. Observational health data sciences and informatics. Available at: https://www.ohdsi.org/data-standardization/the-common-data-model/ .

[32] Siddiqi, N., & House, A. (2006). Delirium: an update on diagnosis, treatment and prevention. Clinical medicine, 6(6), 540.

[33] Lehne, M., Luijten, S., Imbusch, P. V. F. G., & Thun, S. (2019, September). The Use of FHIR in Digital Health-A Review of the Scientific Literature. In GMDS (pp. 52-58).

[34] Devlin, J., Chang, M. W., Lee, K., & Toutanova, K. (2018). Bert: Pre-training of deep bidirectional transformers for language understanding. arXiv preprint arXiv:1810.04805.

## APPENDIX A: ICD codes for disease labeling

**ICD codes for delirium labeling**

- F05: Delirium, not induced by alcohol and other psychoactive substances
    - F05.0: Delirium not superimposed on dementia, so described
    - F05.1: Delirium superimposed on dementia
    - F05.8: Other delirium
    - F05.9: Delirium, unspecified
- F1x.4: Mental and behavioural disorders due to […]. Withdrawal state with delirium.
    - F10.4: Mental and behavioural disorders due to use of alcohol, Withdrawal state with delirium
    - F11.4: Mental and behavioural disorders due to use of opioids, Withdrawal state with delirium
    - F12.4: Mental and behavioural disorders due to use of cannabinoids, Withdrawal state with delirium
    - F13.4: Mental and behavioural disorders due to use of sedatives or hypnotics, Withdrawal state with delirium
    - F14.4: Mental and behavioural disorders due to use of cocaine, Withdrawal state with delirium
    - F15.4: Mental and behavioural disorders due to use of other stimulants, including caffeine, Withdrawal state with delirium
    - F16.4: Mental and behavioural disorders due to use of hallucinogens, Withdrawal state with delirium
    - F17.4: Mental and behavioural disorders due to use of tobacco, Withdrawal state with delirium
    - F18.4: Mental and behavioural disorders due to use of volatile solvents, Withdrawal state with delirium
    - F19.4: Mental and behavioural disorders due to multiple drug use and use of other psychoactive substances, Withdrawal state with delirium
- It is assumed that the F10.4-F19.4 codes do not occur together with F05 codes, as this is part of the ICD code definition.

**ICD codes for sepsis labeling**

- Codes limited to sepsis or close to sepsis, e.g. generalized infection (use these as sepsis labels)
    - A02.1: Salmonella sepsis
    - A20.7: Septicaemic plague
    - A21.7: Generalized tularaemia
    - A22.7: Anthrax sepsis
    - A24.1: Acute and fulminating melioidosis
    - A26.7: Erysipelothrix sepsis
    - A28.0: Pasteurellosis
    - A32.7: Listerial sepsis
    - A33: Tetanus neonatorum <18
    - A39.1: Waterhouse-Friederichsen-Syndrom
    - A39.2: Akute Meningokokkensepsis
    - A39.3: Chronische Meningokokkensepsis
    - A39.4: Meningokokkensepsis, nicht näher bezeichnet
    - A40: Streptococcal sepsis
    - A41: Other sepsis
    - A42.7: Actinomycotic sepsis
    - A48.0: Gasbrand [Gasödem]
    - A48.3: Syndrom des toxischen Schocks
    - B00.7: Disseminated herpesviral disease
    - B37.7: Candidal sepsis
    - B44.7: Disseminierte Aspergillose
    - I33.0: Akute und subakute infektiöse Endokarditis
    - M86.0: Acute haematogenous osteomyelitis
    - O85: Puerperal sepsis
    - O88.3: Obstetric pyaemic and septic embolism
    - P36: Bacterial sepsis of newborn <18
    - R57.2: Septic shock
    - R65: Systemic Inflammatory Response Syndrome
- Codes including, but not limited to sepsis
    - G08: Intracranial and intraspinal phlebitis and thrombophlebitis
    - J95.0: Funktionsstörung eines Tracheostomas
    - O08.0: Infektion des Genitaltraktes und des Beckens nach Abort, Extrauteringravidität und Molenschwangerschaft
    - O75.3: Other infection during labour
    - T81.4: Infektion nach einem Eingriff, anderenorts nicht klassifiziert
    - T88.0: Infection following immunization
- Other codes that may be entered in case of sepsis (still high-risk)
    - A03: Shigellose [Bakterielle Ruhr]
    - A23: Brucellose
    - A28.2 Extraintestinale Yersiniose
    - A39: Meningokokkeninfektion
    - A48.4: Brazilian purpuric fever
    - A54.8: Sonstige Gonokokkeninfektionen
    - H44.0: Purulente Endophthalmitis

- o I40.0: Infective myocarditis
- o O07.0: Misslungene ärztliche Aborteinleitung, kompliziert durch Infektion des Genitaltraktes und des Beckens
- o O07.5: Misslungene sonstige oder nicht näher bezeichnete Aborteinleitung, kompliziert durch Infektion des Genitaltraktes und des Beckens
- o O08.2: Embolie nach Abort, Extrauteringravidität und Molenschwangerschaft
- o T80.2: Infections following infusion, transfusion and therapeutic injection
- o T82.6: Infektion und entzündliche Reaktion durch eine Herzklappenprothese
- o T82.7: Infektion und entzündliche Reaktion durch sonstige Geräte, Implantate oder Transplantate im Herzen und in den Gefäßen
- o T83.5: Infektion und entzündliche Reaktion durch Prothese, Implantat oder Transplantat im Harntrakt
- o T83.6: Infektion und entzündliche Reaktion durch Prothese, Implantat oder Transplantat im Genitaltrakt
- o T84.5: Infektion und entzündliche Reaktion durch eine Gelenkendoprothese
- o T84.6: Infektion und entzündliche Reaktion durch eine interne Osteosynthesevorrichtung [jede Lokalisation]
- o T84.7: Infektion und entzündliche Reaktion durch sonstige orthopädische Endoprothesen, Implantate oder Transplantate
- o T85.7: Infektion und entzündliche Reaktion durch sonstige interne Prothesen, Implantate oder Transplantate
- Other high-risk codes
    - o G00-G09: Inflammatory diseases of the central nervous system
    
    It is assumed that the F10.4-F19.4 codes do not occur together with F05 codes, as this is part of the ICD code definition.

**ICD codes for AKI labeling**

- N14.-: Drug- and heavy-metal-induced tubulo-interstitial and tubular conditions
    - o N14.1: Nephropathy induced by other drugs, medicaments and biological substances
    - o N14.2: Nephropathy induced by unspecified drug, medicament or biological substance
- N17.-: Acute renal failure
    - o N17.0: Acute renal failure with tubular necrosis
    - o N17.1: Acute renal failure with acute cortical necrosis
    - o N17.2: Acute renal failure with medullary necrosis
    - o N17.8: Other acute renal failure
    - o N17.9: Acute renal failure, unspecified
- N19: Unspecified kidney failure
- N99.-: Postprocedural renal failure
- R34: Anuria and oliguria
- R94.4: Abnormal results of kidney function studies

## APPENDIX B: Disease specific leaking features

Leaky features are features that contain information about the label. Including leaky features causes suboptimal training. The following features are removed from the data for each use case:

**Delirium:**

- Any named clinical entity features containing subword DELIR

**Sepsis:**

- Any named clinical entity features containing SEPSIS, SEPTIC, SEPTISCH or SIRS

- Any lab result containing IL_6 or starting with PROCALC

**AKI:**

- Any named clinical entity features matching regular expression:

    NIEREN|ANURIE|FUROSEMID|URINAUSSCHEIDUNG|KARDIO.*RENAL.*SYNDROM|HAEMOFILTRATION|NEPHRITIS|A(C|K)UT.*?NIEREN|ANURIE|ORGAN_FAILURE|(KARDIO|HEPATO).?RENAL.{0,5}SYNDROM|HAEMOFILTRATION|KREA.*|HARNSTOFF

- Any lab result matching regular expression:

    HARNSTOFF|KREA.*|HARNSAEURE|KRISTALLE|PATH.*ZYLINDER|HEFEZELLEN|RUNDEPITHELIEN|MICROALB

There are still potential leaks in medication, which are not yet considered in our leaking feature. This is because currently only two out of the four hospitals provided medication data, both are not coded in ATC and both have limited impact on our prediction models (see Appendix E). We plan to map the medication names to ATC code as our future work, and potential leaking information in medication will also be removed by checking the ATC code.

## APPENDIX C: Excerpt of sample risk assessment in FHIR

```json
{
 "resourceType": "RiskAssessment",
 "id": "case002",
 "contained": [
  {
   "resourceType": "Person",
   "id": "1",
   "gender": "female",
   "birthDate": "1964-01-01"
  },
  {
   "resourceType": "Condition",
   "id": "4",
   "category": [
    {
     "coding": [
      {
       "system": "http://hl7.org/fhir/ValueSet/condition-category",
       "code": "problem-list-item"
      }
     ]
    }
   ],
   "code": {
    "coding": [
     {
      "system": "http://hl7.org/fhir/ValueSet/icd-10",
      "code": "F10.0",
      "display": "F100"
     }
    ]
   },
   "onsetDateTime": "2019-10-21T15:07:22"
  }
```

# APPENDIX D: Characteristics of training datasets

Table 2 lists the characteristics of training datasets of the three use cases in the four hospitals respectively. Each training dataset has a 1:1 balance between records with the targeted disease and those without.

Table 2 Characteristics of training datasets

| Use case | Delirium | | | | Sepsis | | | | AKI | | | |
|---|---|---|---|---|---|---|---|---|---|---|---|---|
| Hospital name | M | H | K | N | M | H | K | N | M | H | K | N |
| Number of records | 19230 | 6456 | 13896 | 46218 | 18600 | 13764 | 8256 | 37704 | 62766 | 33198 | 15582 | 116670 |
| Demographics | | | | | | | | | | | | |
| -Age, median | 61.3 | 66.0 | 66.4 | 64.8 | 60.1 | 64.8 | 65.9 | 63.8 | 60.3 | 65.1 | 65.8 | 64.0 |
| -Female sex, no, (%) | 10492 (55%) | 2062 (32%) | 7082 (51%) | 24243 (53%) | 10343 (56%) | 4424 (32%) | 4151 (50%) | 19663 (52%) | 35246 (56%) | 11064 (33%) | 7887 (51%) | 61393 (53%) |
| Admission type, no, (%) | | | | | | | | | | | | |
| -normal admission | 9900 (52%) | 5768 (89%) | 6095 (44%) | 21147 (46%) | 9635 (52%) | 12233 (89%) | 3542 (43%) | 17087 (45%) | 34221 (55%) | 29539 (89%) | 6826 (44%) | 55359 (47%) |
| -emergency admission | 9249 (48%) | 611 (10%) | 7772 (56%) | 24816 (54%) | 8873 (48%) | 1387 (10%) | 4701 (57%) | 20418 (54%) | 28239 (45%) | 3288 (10%) | 8722 (56%) | 60662 (52%) |
| History of diagnosis, no, (%) | 8819 (46%) | 2556 (40%) | 8072 (58%) | 26435 (57%) | 8372 (45%) | 5527 (40%) | 4806 (58%) | 21429 (57%) | 28152 (45%) | 13366 (40%) | 9190 (59%) | 65013 (56%) |
| Medication, no, (%) | 0 (0%) | 1137 (18%) | 0 (0%) | 2617 (6%) | 0 (0%) | 1919 (14%) | 0 (0%) | 2686 (7%) | 0 (0%) | 4360 (13%) | 0 (0%) | 5834 (5%) |
| Lab Results, no, (%) | 17449 (91%) | 6305 (98%) | 12919 (93%) | 44084 (95%) | 16873 (91%) | 13472 (98%) | 7671 (93%) | 35985 (95%) | 56312 (90%) | 32473 (98%) | 14464 (93%) | 111049 (95%) |
| Vital Sign, no, (%) | 13753 (72%) | 3699 (57%) | 0 (0%) | 18130 (39%) | 13168 (71%) | 7201 (52%) | 0 (0%) | 14911 (40%) | 44541 (71%) | 17262 (52%) | 0 (0%) | 45507 (39%) |
| Named Clinical Entities, no, (%) | 18966 (99%) | 6277 (97%) | 11183 (81%) | 36018 (78%) | 18364 (99%) | 13367 (97%) | 6622 (80%) | 29381 (78%) | 61938 (99%) | 27879 (84%) | 12261 (79%) | 90470 (78%) |

# APPENDIX E: Importance of each feature group

Table 3 evaluates the importance of each feature group in our trained models based on occlusion analysis [27], by checking the impact on the AUROC when a feature group is masked out in the test set. The row of original AUROC without dropping features shows the AUROC of the test set at discharge. Delta to original AUROC after dropping feature shows the gap between the AUROC after masking out each of the indicated feature groups and the original AUROC. A negative value indicates after dropping the feature group in the testing set, the model performs worse. The smaller this delta value is, the greater impact the corresponding features group has on the prediction model. A positive value indicates after dropping this feature group in the test set, the model even performs better, this is likely due to the overfitting during the training.

Table 3. Importance of each feature group

| Use case | Delirium | | | | Sepsis | | | | AKI | | | |
|---|---|---|---|---|---|---|---|---|---|---|---|---|
| Hospital name | M | H | K | N | M | H | K | N | M | H | K | N |
| Original AUROC without dropping feature (%) | 96.74 | 98.03 | 92.57 | 93.09 | 96.48 | 96.17 | 93.91 | 95.55 | 93.41 | 95.75 | 89.18 | 88.09 |
| Delta to original AUROC after dropping feature (%): | | | | | | | | | | | | |
| -Gender | -0.00 | -0.06 | -0.10 | -0.07 | -0.04 | -0.00 | +0.04 | +0.01 | -0.00 | -0.01 | -0.09 | +0.01 |
| -Age group | -0.00 | -0.06 | -0.41 | -0.63 | -0.08 | -0.07 | +0.09 | +0.02 | -0.25 | -0.05 | -0.76 | -0.89 |
| -Admission type | -0.03 | -0.02 | -0.08 | -0.07 | -0.18 | -0.03 | -0.02 | -0.07 | -0.12 | -0.00 | -0.10 | +0.06 |
| -Department of stay | +0.06 | -0.15 | -0.89 | -0.12 | +0.13 | -0.30 | -0.39 | -0.23 | -1.12 | -0.22 | -1.13 | -0.11 |
| -History of diagnosis | -0.18 | -0.11 | -0.91 | -0.62 | -0.12 | -0.25 | -0.46 | +0.07 | +0.04 | -0.18 | -0.65 | -0.22 |
| -Medications | nan | -0.07 | nan | -0.03 | nan | -0.04 | nan | -0.10 | nan | -0.13 | nan | +0.01 |
| -Lab results | -0.90 | -1.17 | -2.19 | -3.38 | -4.39 | -4.07 | -8.86 | -6.38 | -7.55 | -5.39 | -9.21 | -5.30 |
| -Vital signs | -0.01 | -0.03 | nan | -0.31 | -0.04 | -0.20 | nan | -0.08 | +0.11 | -0.04 | nan | -0.00 |
| -Named clinical entities | -13.85 | -4.23 | -7.82 | -7.93 | -2.32 | -1.64 | -1.99 | -1.07 | -1.83 | -0.25 | -0.42 | -0.86 |

## APPENDIX F: Impact of data augmentation and model calibration

Table 4 shows the impact of our data augmentation policy and calibration by comparing the performance of several delirium risk prediction models at HDZ. The baseline is a model calibrated in HDZ following the scalable model development process with data augmentation applied, its metrics are also presented in Table 1. The second model is a model calibrated in HDZ, but without applying data augmentation. The third model is a model trained at the development site, and evaluated at HDZ without model calibration.

Table 4. Impact of data augmentation and model calibration

| Delirium use case at HDZ | Baseline model (with calibration, with data augmentation) | Model without data augmentation | Model developed at development site |
|---|---|---|---|
| admission AUROC (CI 95%) | 75.20% [74.7 - 75.7] | 71.30% [70.8 - 71.8] | 61.84% [61.3 - 62.4] |
| discharge AUROC (CI 95%) | 98.03% [97.9 - 98.2] | 97.50% [97.3 - 97.7] | 93.84% [93.6 - 94.1] |

# APPENDIX G: Model architecture and hyperparameters

The prediction models are trained with the same model architecture: we use Transformer (Tensor2Tensor) [24] to train a binary classification model for clinical risk prediction. Our prediction model is built with the transformer_small model with text2class problem. The encoder is built with two transformer layers, with hidden size 256, and 4 attention heads. Since we directly generate the predicted binary class as the only output, there is no decoder required. We build our own vocabulary, which consists of the feature values from the records in the training set. During the prediction, e.g. on the test set, those feature values that are not contained in our vocabulary are discarded. We use Adagrad as the optimizer, and we train our prediction models with 80000 training steps, roughly 2 epochs; the number of steps are slightly adjusted based on the number of training samples in different use cases.